\documentclass[letterpaper, 10 pt, conference]{ieeeconf} 
\IEEEoverridecommandlockouts
\usepackage{graphicx}
\usepackage{multirow}
\usepackage{pbox}
\usepackage{xcolor}
\usepackage[super]{nth}
\usepackage{amsmath}
\usepackage{tikz}
\usepackage{textcomp}
\usepackage{url}
\usepackage{hyperref}
\usepackage[hyphenbreaks]{breakurl}

\usepackage{xspace}
\usepackage{float}
\usepackage{afterpage}
\usepackage{pgfplots}

\usepackage{footnote}
\makesavenoteenv{tabular}

\newcommand{\eg}{e.g.\@\xspace}

\newcommand{\etal}{et al.\@\xspace}

\pgfplotsset{compat=1.18}

\title{\LARGE \bf
The Autonomous Software Stack of the FRED-003C:\\The Development That Led to Full-Scale Autonomous Racing
}

\author{Zalán Demeter, Levente Puskás, Balázs Kovács, Ádám Matkovics, Martin Nádas, Balázs Tuba,\\ Zsolt Farkas, Ármin Bogár-Németh, Gergely Bári%
\thanks{BME Formula Racing Team, Hungary (url: \url{https://frtbme.hu/})}%
\thanks{Széchenyi István University, H-9026 Győr, Hungary (e-mail: zalan.demeter@humda.hu)}%
}

\newcommand\copyrighttext{%
  \footnotesize \textcopyright 2025 IEEE. Personal use of this material is permitted. Permission from IEEE must be obtained for all other uses, in any current or future media, including reprinting/republishing this material for advertising or promotional purposes, creating new collective works, for resale or redistribution to servers or lists, or reuse of any copyrighted component of this work in other works.
  }
\newcommand\copyrightnotice{%
\begin{tikzpicture}[remember picture,overlay]
\node[anchor=south,yshift=10pt] at (current page.south) {\fbox{\parbox{\dimexpr\textwidth-\fboxsep-\fboxrule\relax}{\copyrighttext}}};
\end{tikzpicture}%
}

\begin{document}

\maketitle
\copyrightnotice
\thispagestyle{empty}
\pagestyle{empty}

\begin{abstract}

Scientific development often takes place in the context of research projects carried out by dedicated students during their time at university. In the field of self-driving software research, the Formula Student Driverless competitions are an excellent platform to promote research and attract young engineers. This article presents the software stack developed by BME Formula Racing Team, that formed the foundation of the development that ultimately led us to full-scale autonomous racing. The experience we gained here contributes greatly to our successful participation in the Abu Dhabi Autonomous Racing League. We therefore think it is important to share the system we used, providing a valuable starting point for other ambitious students. We provide a detailed description of the software pipeline we used, including a brief description of the hardware-software architecture. Furthermore, we introduce the methods that we developed for the modules that implement perception; localisation and mapping, planning, and control tasks.

\end{abstract}

\section{Introduction}
\label{sec:introduction}

Innovation and hands-on effort by aspiring engineers have been fueling rapid development in the sector of autonomous vehicle technology, with many getting foundational experience by competing in events such as Formula Student (FS). Events like these challenge teams not only to design and build autonomous systems but also cultivate the skills necessary to tackle larger-scale problems in the field.

This article describes the autonomous software stack and architecture developed by BME Formula Racing Team. This project is the result of years of collaborative research and iterative development of university students that focus on competing in FS races. The significance of such efforts goes beyond academic competitions. The skills developed during the project have since translated into very important contributions to the state-of-the-art of full-scale autonomous racing such as those showcased in the Abu Dhabi Autonomous Racing League (A2RL). By sharing the methods and experiences of the FRED-003C project, we hope to inspire the next generation of engineers and demonstrate how student initiatives play a key role in promoting STEM education and innovation.

    \begin{figure}[!ht]
     \centering
     \includegraphics[width=8.0cm]{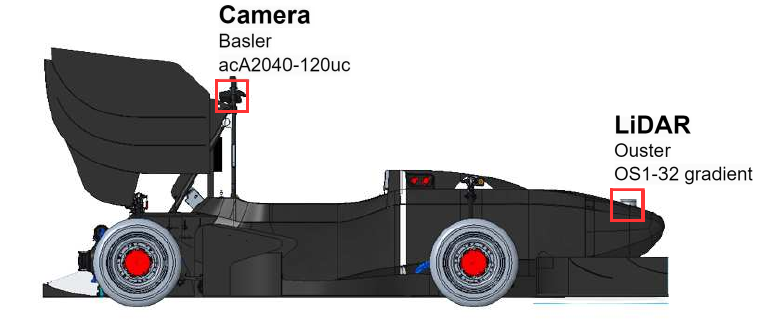}
     \caption{The perception sensor placement of the FRED-003C. The LiDAR sensor is placed on the nose cone and the camera equipped with a wide angle lens is mounted to the top of the main hoop.}
      \label{fig:sensor-placement}
    \end{figure}

    \begin{figure*}
     \centering
     \includegraphics[width=16.0cm]{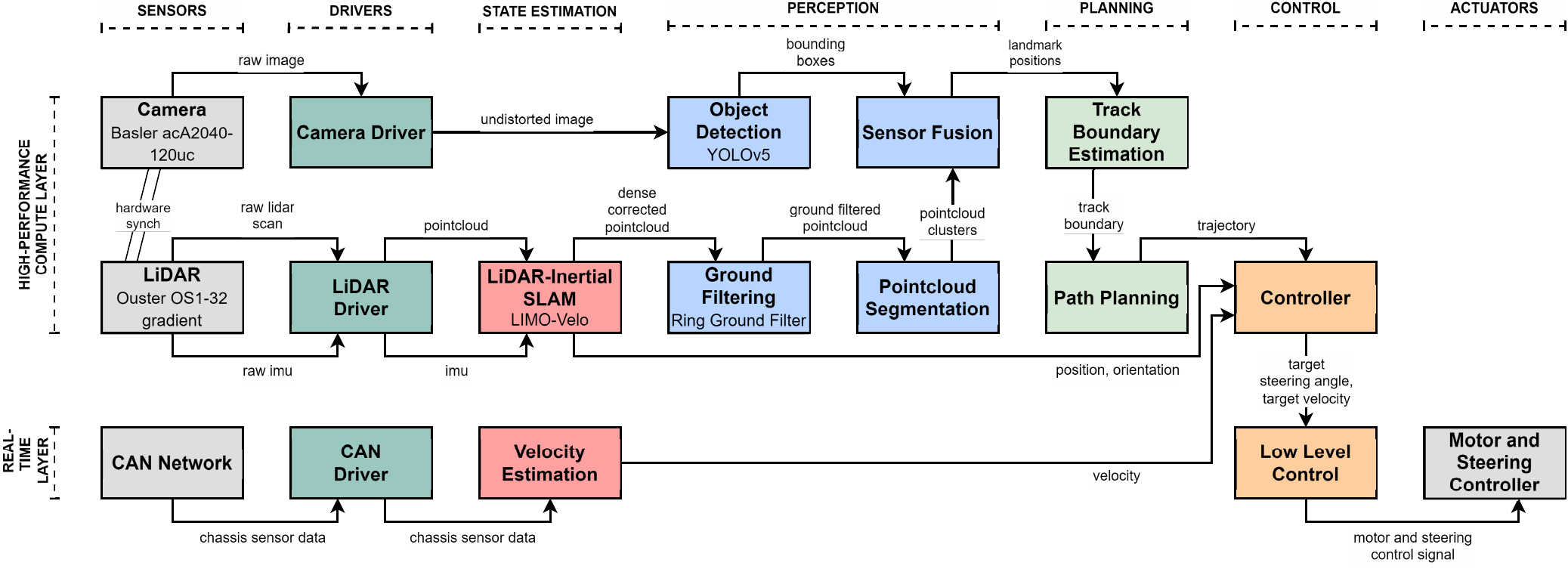}
     \caption{Block diagram depicting the hardware-software architecture of the autonomous system. The software pipeline can be divided into four parts; state estimation, perception, planning, and control. The execution is distributed between a high-performance compute layer and a real-time layer. }
      \label{fig:arch}
    \end{figure*}

\section{Literature Review}
\label{sec:literature-review}

In the world of autonomous racing, the competitions are held on different scales with different vehicle platforms and race rules. This results in a wide range of research findings on a variety of diverse topics. This allows a large number of scientific studies to be used as a source of inspiration for the implementation of such a project. These include research conducted for F1Tenth \cite{OKelly2019}, Formula Student Driverless (FSD) \cite{amz2019}, or even full-scale series such as the Indy Autonomous Challenge \cite{Raji_2024}. Throughout \hyperref[sec:methods]{Section \ref{sec:methods}}, a wide range of articles will be referenced and their significance will be discussed. Here, for the sake of brevity, we briefly mention only the two publications that provide the most important basis of comparison. For a more comprehensive understanding, the reader is encouraged to familiarise themselves with the rules of FSD, as detailed in the official rulebook \cite{fsgrules}.

The article by Kabzan \etal \cite{amz2019} is one of the first to present a complete FSD software stack and is still of great relevance. This architecture presented by AMZ Racing follows a pipeline structure and its modules can be partitioned according to the sense-think-act paradigm \cite{1218700}, which is a widely adopted approach. The other article we mention is that of Alvarez \etal \cite{ka-raceing}, which presents the award-winning software stack used by KA-RaceIng. This is of high importance because they are presenting a supervisory architecture which contributes greatly to the robust and error-free operation of their system.

\section{Methods}
\label{sec:methods}

The vehicle platform used to develop the self-driving software stack varies annually, with iteratively updated versions used to test and refine the driverless pipeline. \hyperref[fig:sensor-placement]{Figure \ref{fig:sensor-placement}} shows the vehicle code named FRED-003C, the latest in this development cycle. The diagram also shows the environment perception sensors used for the Autonomous System (AS) and their placement. The purpose of these sensors is to enable the detection of cones that bound the track in the widest possible field of view (FoV) and distance. However, the selection of sensors and the decision process for their placement is beyond the scope of this paper.

The proposed software architecture can be divided into four logical subsystems; state estimation, perception, planning, and control. The state estimation is responsible for extracting important states related to the vehicle, either through measurement or estimation procedures. The modules in perception preprocess the data from the corresponding sensors and interpret the environmental information. The task of planning is to devise an optimal trajectory within the currently traversable area. The algorithms involved in control generate safe and enforceable control signals for the vehicle based on given references. \hyperref[fig:arch]{Figure \ref{fig:arch}} shows a block diagram of the software pipeline that implements this division. The diagram also shows the hardware components used and the software drivers that interface with them. The execution of software units in the AS is implemented on two layers, which can also be seen in this figure; the high-performance compute layer and the real-time layer.

\subsection{State Estimation}
\label{sec:state-estimation}

State estimation is a key component of the pipeline, as all other modules depend on it, and its accuracy directly impacts their performance. In the FSD competition, the layout of the autocross and trackdrive event is not known beforehand, which further complicates this problem, as to be able to localise on the track, first a map needs to be built. This raises the Simultaneous Localisation And Mapping (SLAM) problem. A possible approach without SLAM is to perform local estimation from current measurements only. This approach presents significant limitations, including susceptibility to perception failures and the inability to perform global optimisations.

\subsubsection{Fast SLAM}
To overcome these limitations, we implemented an algorithm similar to FastSLAM \cite{fastslam}, which is an accessible and straightforward landmark-based SLAM algorithm. It is based on the principle that localisation and mapping parts are conditionally independent and thus uses particle filtering for localisation coupled with an Extended Kalman Filter (EKF) to estimate landmark positions. Each particle contains a position and map estimate. A weight is assigned to each particle according to their alignment with the current measurement. Based on the weight of particles, at every iteration resampling selects the highest weighing particle to output the corresponding map and pose.

FastSLAM 2.0 \cite{fastslam2} addresses some limitations in accuracy and robustness by incorporating sensor measurements into the proposal distribution, which is found to be particularly effective in cases where motion noise exceeds measurement noise. These modifications were later adopted into our implementation.

\subsubsection{Graph SLAM}

Previous experiments with graph-based algorithms encountered performance limitations, preventing real-time execution. To address this, we integrated a graph optimisation framework with FastSLAM, which serves as a front-end for landmark and pose estimation while concurrently building and optimising a pose graph during loop closures. However, poor data associations can affect optimisation performance, which FastSLAM resolves by using the highest-weight particle that best fits the measurements.

For the pose graph optimisation, we propose constructing a graph with landmarks and poses along the path and minimising two residuals: the landmark observation error and the pose motion error to optimise both trajectory and landmark positions \cite{Viragh2022}.

Given a robot pose \( \mathbf{p} = [p_x, p_y, p_\theta]^\top \), a landmark position \( \mathbf{l} = [l_x, l_y]^\top \), and a measurement \( [m_r, m_\theta] \) representing range and bearing, the landmark observation residual is:
\begin{equation}
\mathbf{r}_{\text{landmark}} = 
\begin{bmatrix}
\cos(p_\theta) & \sin(p_\theta) \\
-\sin(p_\theta) & \cos(p_\theta)
\end{bmatrix}
\begin{bmatrix}
l_x - p_x \\
l_y - p_y
\end{bmatrix}
-
\begin{bmatrix}
m_r \cos(m_\theta) \\
m_r \sin(m_\theta)
\end{bmatrix}
\end{equation}

For consecutive poses \( \mathbf{p}_{\text{prev}} = [p_{x,\text{prev}}, p_{y,\text{prev}}, p_{\theta,\text{prev}}]^\top \) and \( \mathbf{p}_{\text{curr}} = [p_{x,\text{curr}}, p_{y,\text{curr}}, p_{\theta,\text{curr}}]^\top \), the pose motion residual is defined using velocity \( v \) and angular velocity \( \omega \) as follows:
\begin{equation}
\mathbf{r}_{\text{pose}} = 
\begin{bmatrix}
(p_{x,\text{curr}} - p_{x,\text{prev}}) - v \cos(p_{\theta,\text{prev}} + \omega) \\
(p_{y,\text{curr}} - p_{y,\text{prev}}) - v \sin(p_{\theta,\text{prev}} + \omega) \\
(p_{\theta,\text{curr}} - p_{\theta,\text{prev}}) - 2\omega
\end{bmatrix}
\end{equation}
where \( v \) is the linear velocity and \( \omega \) is the angular velocity associated with the motion between poses.

\subsubsection{LiDAR SLAM}

We reached the performance limitations of our previous approach and had to transition to a more state-of-the-art solution. Today, LiDAR-based and especially LiDAR-inertal algorithms are widely adopted in the field of SLAM. Some notable research in this field was done by Xu \etal \cite{fastlio2}, who developed a fast, robust, and versatile LiDAR-inertial odometry framework. The approach we used was based on FAST\_LIO2 but was adapted to work on the dynamic, fast direction-changing motion of a Formula Student racecar, developed by Segarra \etal \cite{limo_velo} at UPC Barcelona.

\subsection{Perception}
\label{sec:perception}

The primary function of the perception system is to provide landmarks for the planner. The system must ensure the precise detection of environmental features to facilitate reliable localisation, mapping, and planning, as erroneous data can cause the system to demonstrate undefined behaviour. Therefore, we propose a standalone LiDAR perception system which performs independent object detection and environmental mapping, significantly improving the system's robustness against adverse or variable environmental conditions. This is complemented by fusing information from camera detections to facilitate colour classification and further improve the robustness of the sensing. 

The real-time operation of the system requires precise timing, achieved by hardware synchronisation between the camera and the LiDAR sensor, which ensures alignment of images and point clouds. In this arrangement the LiDAR acts as a trigger source for the camera and triggers it precisely at the same position once every rotation. This ensures that the images are captured at consistent and predictable intervals, aligned with the LiDAR's rotational movement.

\subsubsection{Camera Perception}
\label{sec:camera-perception}

\begin{figure}[!ht]
     \centering
     \includegraphics[width=8.0cm]{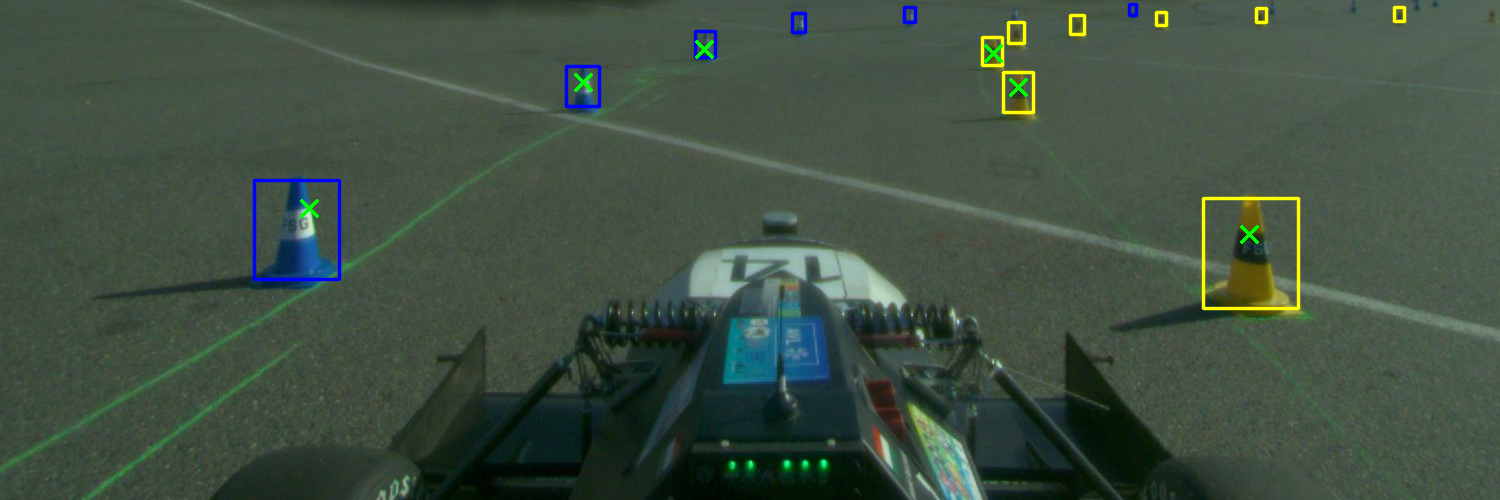}
     \caption{LiDAR cluster centers marked by green crosses are projected onto the image containing bounding boxes of the detected cones.}
      \label{fig:perception}
    \end{figure}

The pipeline starts with the camera driver, which performs low-level preprocessing steps such as lens undistortion and region-of-interest segmentation. The frames are then processed using the YOLOv5 neural network \cite{yolov5}, a state-of-the-art object detection algorithm for real-time applications. The network has been trained on the FSOCO dataset \cite{fsoco_2022}, which contains images of cones in different environments, helping the model to generalise well in a competition setting. The detected objects are classified into four categories: blue cones, yellow cones, small orange cones, and large orange cones shown in \hyperref[fig:perception]{Figure \ref{fig:perception}}.

\subsubsection{Lidar Perception}
\label{sec:lidar-perception}

The pipeline starts with the LiDAR driver, which accumulates data between scans to retain all relevant information for further processing. Given the use of a rotating LiDAR, motion-induced distortions may occur during data capture. To address this, a motion correction algorithm is applied to the point cloud, adjusting each point to its precise location in space. These two steps are integral to our LiDAR SLAM system, but earlier we used an independent LiDAR correction solution based on the paper of Renzler \etal \cite{distortion_correction}. The final preprocessing step involves the removal of the ground plane using a slope-robust cascaded ground segmentation technique \cite{ground_filtering}. 

    \begin{figure}[!ht]
     \centering
     \includegraphics[width=8.0cm]{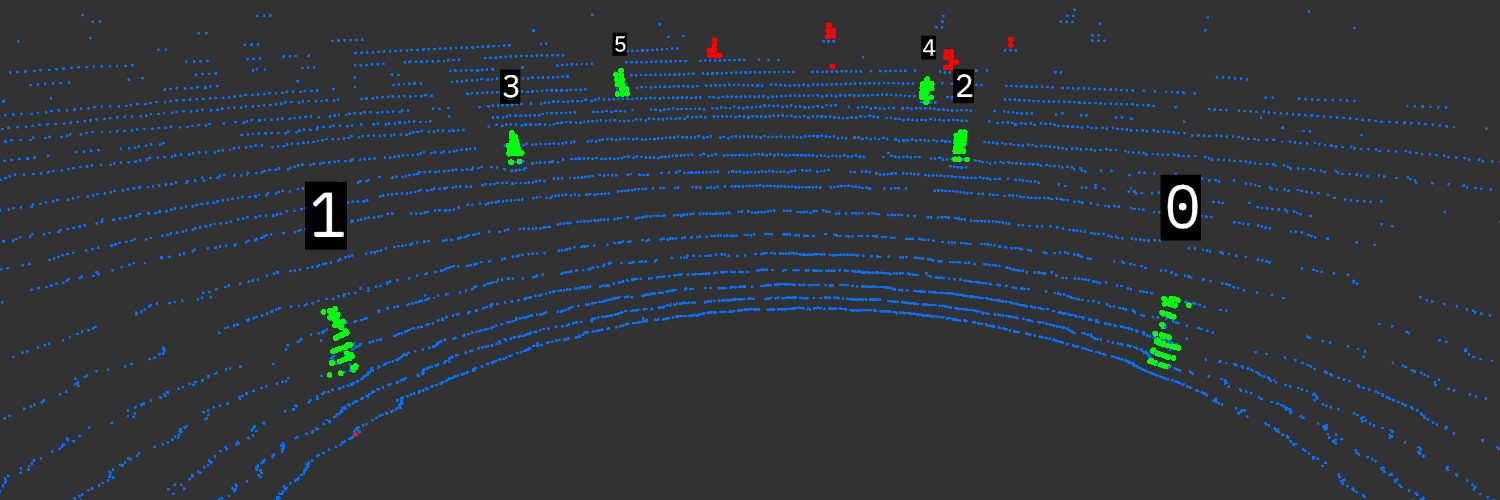}
     \caption{LiDAR point cloud segmented into ground points marked by blue, non ground points marked by red, and classified cones marked by green.}
      \label{fig:lidar_perception}
    \end{figure}

After preprocessing, cone detection is carried out on the corrected point cloud using Euclidean distance-based clustering shown in \hyperref[fig:lidar_perception]{Figure \ref{fig:lidar_perception}}. The ground filtering process typically removes a significant number of points near the base of the cones. To address this, a cone reconstruction step fits a cylinder around the cluster's central axis, re-adding previously removed points. Following clustering, rule-based filtering is applied, evaluating the size, shape, symmetry, and point density of each cluster. These filters enable the elimination of non-cone objects. 

For cone colour classification, intensity readings from the LiDAR are used to distinguish blue and yellow cones based on the intensity profile of their stripes. For blue cones, intensity values peak in the middle, whereas for yellow cones, intensity readings dip in the middle due to the black stripe. After segmenting the point cloud into layers, we calculate the median intensity of each layer and fit a second-degree polynomial to the values. A positive slope in the fitted curve indicates a blue cone, while a negative slope indicates a yellow cone.

\subsubsection{Sensor Fusion}
This module integrates LiDAR and camera data to provide accurate cone location and colour information for the path planner. The cluster centres of the cones detected by the lidar pipeline are projected onto the image plane, shown in \hyperref[fig:perception]{Figure \ref{fig:perception}}, using a projective transformation
\begin{equation}
\begin{bmatrix} x \\ y \\ w \end{bmatrix} = \underbrace{\begin{bmatrix} f_x & 0 & c_x \\ 0 & f_y & c_y \\ 0 & 0 & 1 \end{bmatrix}}_{\text{K}} \cdot \underbrace{\begin{bmatrix} r_{11} & r_{12} & r_{13} & t_x \\ r_{21} & r_{22} & r_{23} & t_y \\ r_{31} & r_{32} & r_{33} & t_z \end{bmatrix}}_{\text{[R$|$T]}} \cdot \begin{bmatrix} X \\ Y \\ Z \\ 1 \end{bmatrix}
\end{equation}
where $K$ is the intrinsic calibration matrix, $[R|T]$ is the extrinsic calibration matrix, $[x, y, w]^T$ is the image point, and $[X, Y, Z, 1]^T$ is the world point. If the projected point lies outside the camera’s FoV, LiDAR data is used exclusively to allow for wider overall FoV. If it falls within the camera’s FoV, further processing determines if it lies inside or outside the detected bounding boxes. Points inside the boxes are classified with the camera’s colour data, while points outside are considered false detections. This method refines cone detection by reducing reliance on camera and improves cone classification through a combination of both sensor modalities. The final output includes the 3D coordinates and colour of the detected cones.

\subsection{Planning}
\label{sec:planning}
    
\begin{figure}[!ht]
            \centering
            \begin{minipage}{4cm}
                \centering
                \includegraphics[height=2.75cm]{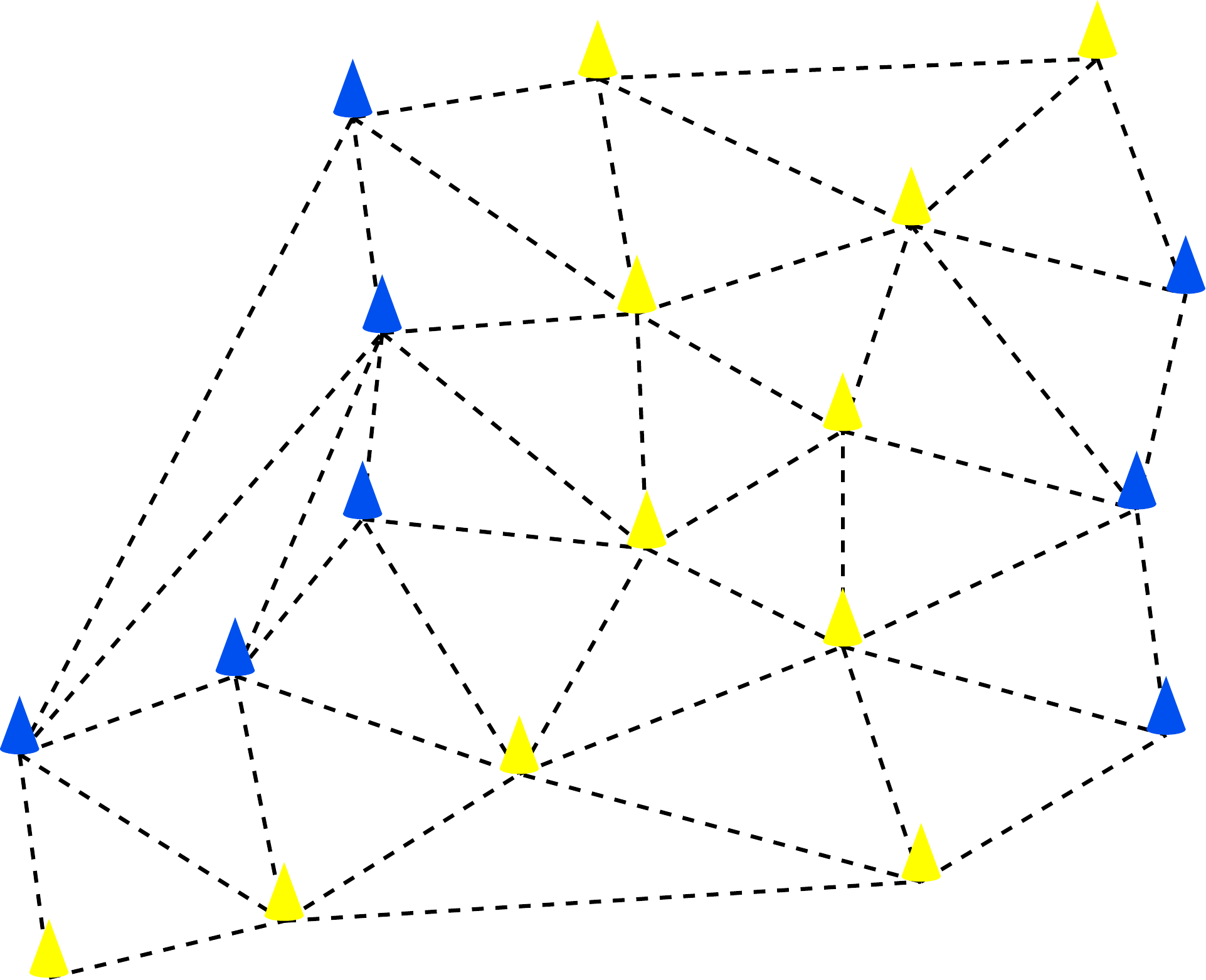}
                \caption{Delaunay triangulation, where yellow and blue triangles represent the cones that bound the track, and the dashed lines show the resulting partitioning.}
                \label{fig:delaunay}
            \end{minipage}
            \hfill
            \begin{minipage}{4cm}
                \centering
                \includegraphics[height=2.75cm]{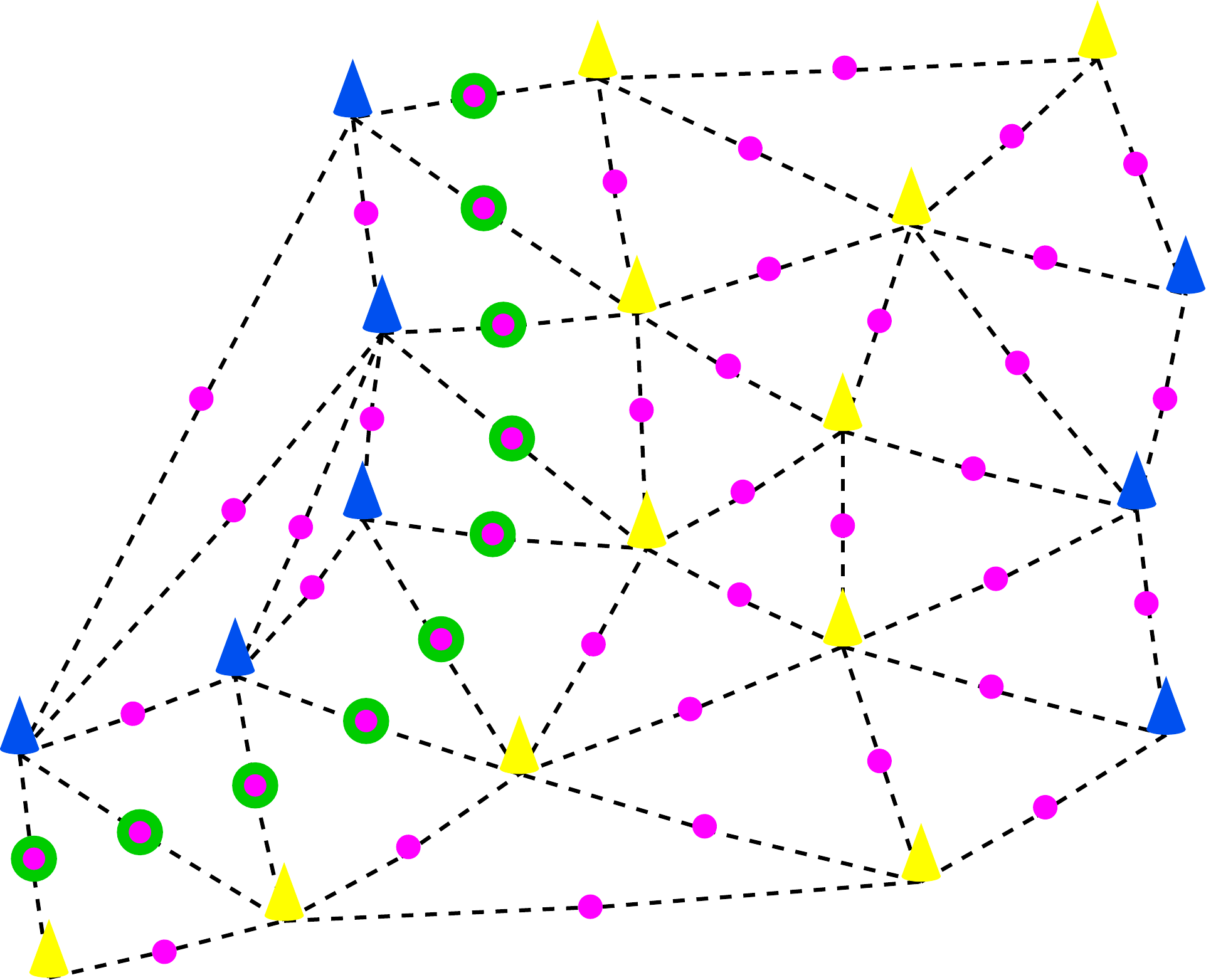}
                \caption{Centreline search, where pink dots mark the midpoints of triangle edges, and the green dots represent the centerline points of the desired trajectory.}
                \label{fig:centreline_search}
            \end{minipage}
        \end{figure}

    The proposed path planner generates a centerline trajectory and the corresponding velocity profile based on the dynamic constraints of the vehicle, given a 2D map of coloured cones. The choice of the applied method depends on the specific dynamic event that is addressed. During the skidpad event, a precalculated path is loaded and aligned with the observed cones using a Generalised Iterative Closest Point algorithm \cite{gicp}. Subsequently, a velocity profile fitting algorithm is applied to the path, resulting in the final reference. For all other events, Delaunay triangulation is used on the cone map in which the centerline of the track runs through the midpoints of the triangle edges, thus the centreline search algorithm is applied to these midpoints. This is illustrated in \hyperref[fig:delaunay]{Figure \ref{fig:delaunay}} and \hyperref[fig:centreline_search]{Figure \ref{fig:centreline_search}}. 

    The proposed centreline search algorithm works on these midpoints -- marked pink in the figure -- and introduces a reward function-based approach to determining successive centreline points. This method evaluates potential neighbouring points by applying a reward function that integrates several key factors derived from Formula Student regulations, including the colour of surrounding cones, track width, changes in angle, and the distance between midpoints. In addition, the algorithm integrates an event horizon to evaluate all feasible future paths and also uses a predictive model informed by previously determined centerline points. This predictive model estimates the position of the next midline point based on the tendencies observed in the previous midpoints, changes in angle, and distances between the points. Midpoints that are closer to the predicted position are assigned higher rewards. Ultimately, the algorithm selects the midpoints that yield the highest cumulative reward value.

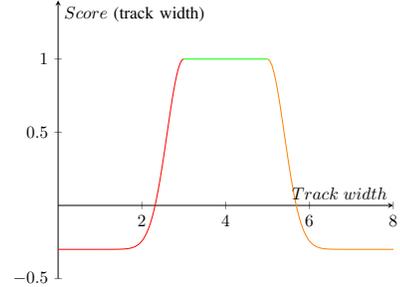
\begin{figure}[!ht]
        \centering
        \begin{tikzpicture}[scale=0.65]
            \begin{axis}[
                axis lines = middle,
                xlabel = $Track\;width$,
                ylabel = {$Score{\text{ (track width)}}$},
                domain=0:10,
                samples=200,
                legend pos=south west,
                xmin=0, xmax=8, ymin=-0.5, ymax=1.4
            ]
    
            \addplot [
                domain=0:3, 
                samples=100, 
                color=red,
            ] {1.3* exp(-((x-3)/0.4)^2 / 2) - 0.3};
    
            \addplot [
                domain=3:5, 
                samples=2, 
                color=green,
            ] {1};
    
            \addplot [
                domain=5:10, 
                samples=100, 
                color=orange,
            ] {1.3* exp(-((x-5)/0.4)^2 / 2) - 0.3};
    
            \end{axis}
        \end{tikzpicture}
        \caption{The reward function of the track width property for a midpoint, where function characteristics are designed to enforce competition rules.}
        \label{fig:reward_function}
    \end{figure}

   At each midpoint, a score is calculated for every reward factor, which is then aggregated to determine the final reward value. For example, the track width is one of the factors evaluated with an optimal range of 3 to 5 metres specified by Formula Student rules. The dedicated reward function -- shown in \hyperref[fig:reward_function]{Figure \ref{fig:reward_function}} -- for the track width promotes midpoints that fall within this optimal range while penalising those that deviate significantly.
    
    Several potential inaccuracies may exist in the cone map due to factors such as the 2-3 cm uncertainty of the lidar sensor, processing noise (\eg in the motion correction or SLAM algorithm) and human error in cone placement by track marshals. In order to account for these inaccuracies, the transition between promotion and penalisation is not stepwise, rather a Gaussian function is used as a smoothing term, as shown in \hyperref[eq:track_width]{Equation \ref{eq:track_width}}

    \begin{equation}
        f(x) =
        \begin{cases} 
        1.3 \cdot \exp\left(-\frac{\left(\frac{x-3}{0.4}\right)^2}{2}\right) - 0.3, & \text{if } x < 3 \\
        1, & \text{if } 3 \leq x \leq 5 \\
        1.3 \cdot \exp\left(-\frac{\left(\frac{x-3}{0.4}\right)^2}{2}\right) - 0.3, & \text{if } x > 5
        \end{cases}
        \label{eq:track_width}
    \end{equation}
    
    Certain reward factors are prioritised over others (\eg track width $\gg$ prediction), and each factor is assigned a weight to enhance its influence. The final score for each midpoint is calculated as the weighted sum of all reward factors:
    \begin{equation}
     \text{Reward}_{\text{midpoint}} = \sum_{i \in \text{Reward Factors}} \text{Weight}_i \cdot \text{Score}_i
    \end{equation}
    Then the calculated rewards are compared and the midpoint with the highest value is selected.
    
    The complexity of this cost function ensures robustness against multiple simultaneous perception or SLAM related errors. Furthermore, the algorithm can function without reliance on cone colour information, enabling lidar-only sensor configurations to accurately plan a feasible trajectory.
        
\begin{figure}[!ht]
        \centering
        \begin{minipage}{2.666cm}
            \centering
            \includegraphics[height=3.0cm]{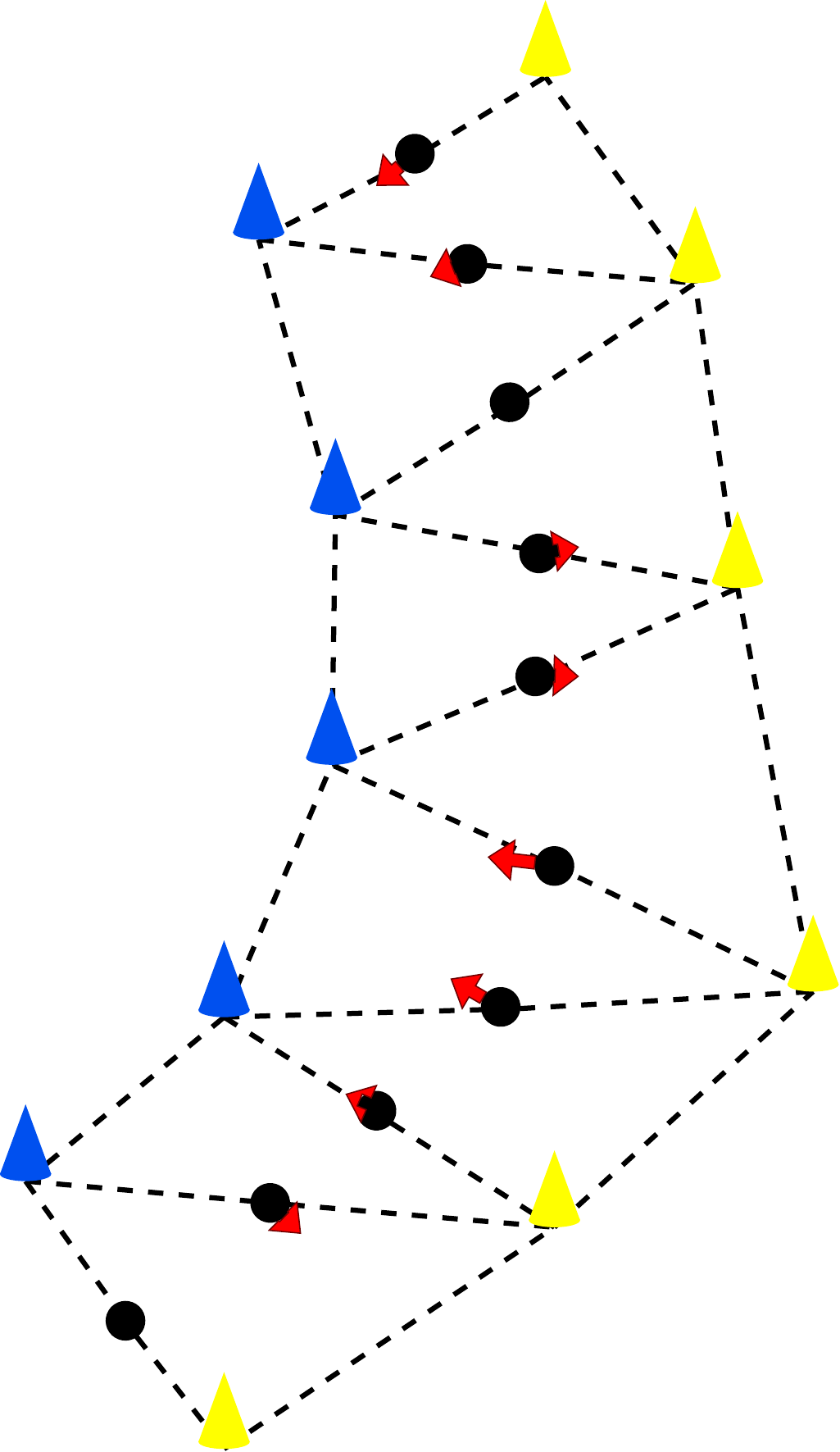}
            \caption{Combined smoothing aligns center points (black dots) with given vectors (red arrows) to minimize curvature variation.}
            \label{fig:smoothing}
        \end{minipage}
        \hfill
        \begin{minipage}{2.666cm}
            \centering
            \includegraphics[height=3.0cm]{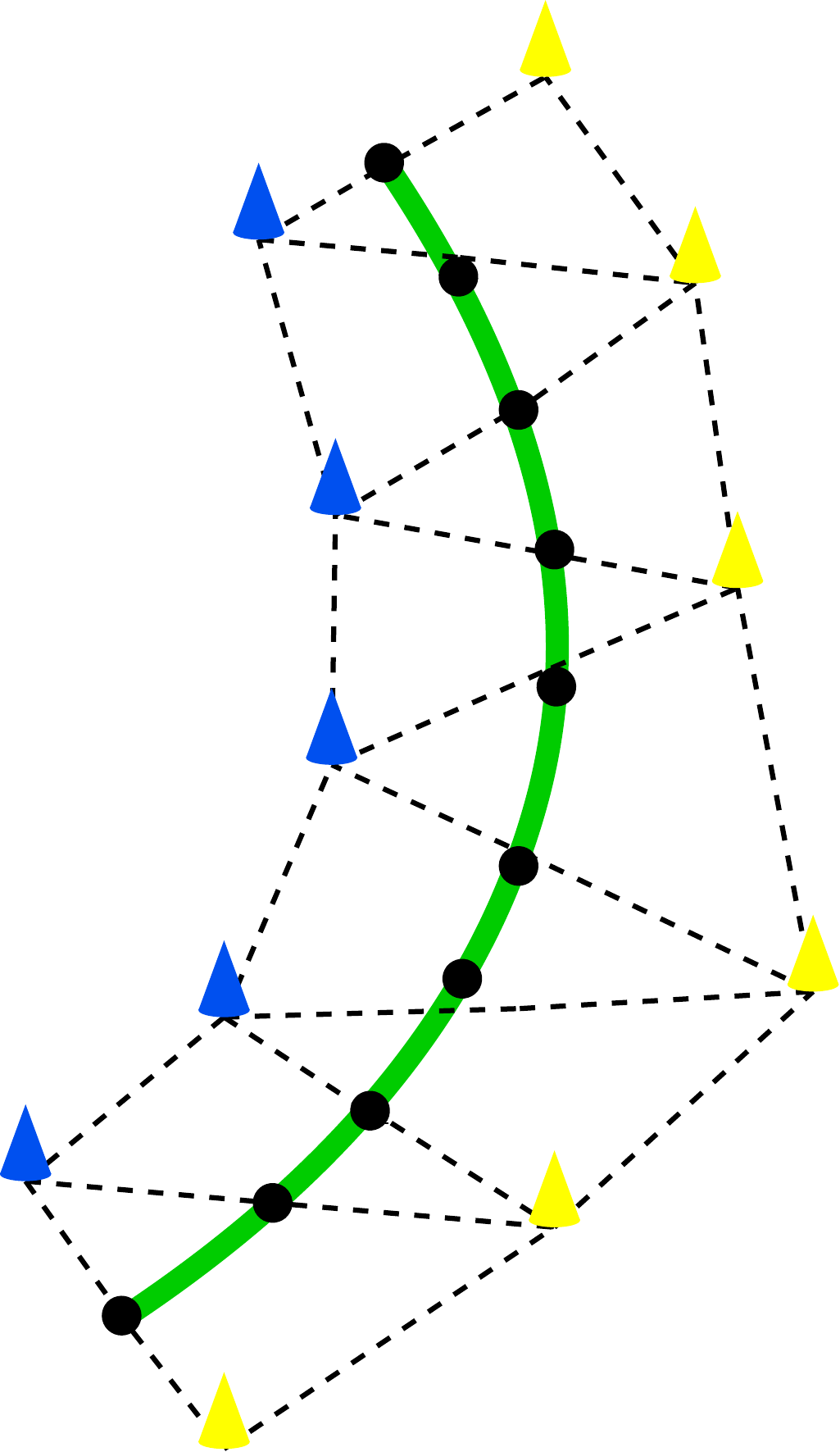}
            \caption{For nonlinear paths $C^2$ spline fitting is used to devise a reference trajectory marked by the green curve.}
            \label{fig:spline_fitting}
        \end{minipage}
        \hfill
        \begin{minipage}{2.666cm}
            \centering
            \includegraphics[height=3.0cm]{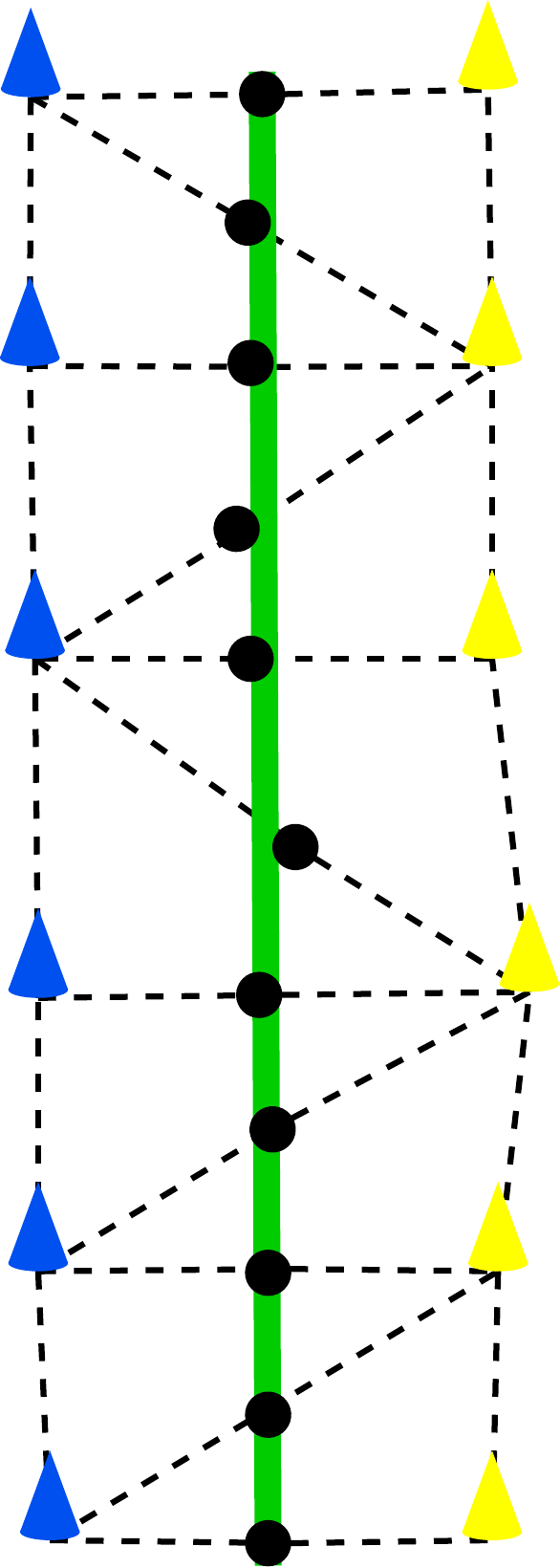}
            \caption{For linear paths a best-fit algorithm is used to devise a reference trajectory marked by the green line.}
            \label{fig:line_fitting}
        \end{minipage}
    \end{figure}

    The centreline search works in a greedy way and initiates from the vehicle's position, applying the cost function iteratively to identify the next centreline point. Once all relevant centreline points are established ahead of the vehicle, smoothing techniques are applied to eliminate outliers and achieve a lower average curvature path, which can be seen in \hyperref[fig:smoothing]{Figure \ref{fig:smoothing}}. This involves a combination of moving average smoothing and Opheim simplification. \hyperref[tab:smoothing_comparison]{Table \ref{tab:smoothing_comparison}} provides a quantitative analysis of smoothing methods, using key metrics such as lap time, lateral acceleration, curvature variation rate (CVR), and minimum distance to boundaries. The lap time is derived using the velocity profile fitting method described later, and to isolate the influence of the path geometry from the velocity-related constraints, the lateral acceleration values are computed assuming a constant velocity of $5 m/s$. The combined approach yields the lowest lap time, overall CVR, and standard deviation in lateral acceleration, while ensuring a safe distance from boundaries, indicating a smoother and more dynamically efficient trajectory.
    
    \begin{table}[ht]
        \centering
        \setlength{\tabcolsep}{4pt}
        
        \caption{Comparison of trajectory smoothing methods on the track used in the 2023 Formula Student Germany competition}

        \begin{tabular}{lcccc}
            \hline
            & \textbf{Raw} & \textbf{Moving Avg.} & \textbf{Opheim} & \textbf{Combined} \\
            \hline
            Lap time [s] & 58.579 & 39.014 & 41.161 & 38.438 \\
            Max lat. acc [m/s²] & 40.138 & 12.325 & 11.115 & 13.502 \\
            Mean lat. acc [m/s²] & 2.980 & 1.660 & 1.674 & 1.562 \\
            Std lat. acc [m/s²] & 3.839 & 1.556 & 1.590 & 1.419 \\
            Max CVR [$1/\text{m}^2$] & 4.185 & 1.263 & 1.309 & 0.079 \\
            Mean CVR [$1/\text{m}^2$] & 0.398 & 0.036 & 0.113 & 0.002 \\
            Std CVR [$1/\text{m}^2$] & 0.496 & 0.086 & 0.136 & 0.003 \\
            Min dist. to bound [m] & 1.547 & 1.227 & 1.349 & 1.261 \\
            \hline
        \end{tabular}
    
        \label{tab:smoothing_comparison}
    \end{table}
    
    Following the smoothing process, line fitting is used for acceleration shown in \hyperref[fig:line_fitting]{Figure \ref{fig:line_fitting}}, and cubic spline fitting is used for autocross and track drive events shown in \hyperref[fig:spline_fitting]{Figure \ref{fig:spline_fitting}}. The cubic spline is preferred because of its twice-continuously differentiable ($C^2$) property, allowing for the calculation of orientation and curvature values at each point while meeting the controller requirements. The smoothing of centreline points and the distance between them prevent large divergences during spline fitting.

    Using orientation and curvature, a modified forward-backward solver is used to calculate a velocity profile \cite{velocity_profile_fitting} that results in laptime-optimal scalar velocities at each point under the dynamic constraints of the vehicle. The algorithm has been adjusted to acknowledge that the vehicle may not achieve maximum acceleration and deceleration in all scenarios (\eg during turns), ensuring that only dynamically feasible velocities are requested. This concludes the generation of the reference trajectory that includes x- and y-coordinates, orientation, curvature, and scalar velocity information with a defined sampling rate.

\subsection{Control}
\label{sec:control}

    Similarly to driver action, our controller's outputs are the steering angle and the throttle. Our approach consists of distinct lateral and longitudinal control schemes which remain reliable for both low and high velocities, unlike dynamic model-based controllers, which do not accurately capture vehicle behaviour at lower speeds. This is important given the typical layout of Formula Student tracks, where low-speed maneuverability is critical. In the article by Kabzan \etal \cite{amz2019} this limitation is addressed by combining the solutions of two separate Model Predictive Controllers; our approach achieves this without the need for additional complexity. Furthermore, our method has only a few straightforward-to-tune parameters and does not rely on access to additional software such as fast optimisation solvers. This approach is particularly well-suited when testing time is limited, there is no access to sufficiently accurate simulations for tuning control parameters, and there is a lack of observability or sensors -- such as those for estimating lateral velocity or vehicle parameters like inertia and complex tyre characteristics -- that would otherwise be necessary. 
    
    \subsubsection{Lateral Control}
    
\begin{figure}[!ht]
            \centering
            \begin{minipage}{4cm}
                \centering
                \includegraphics[width=4cm]{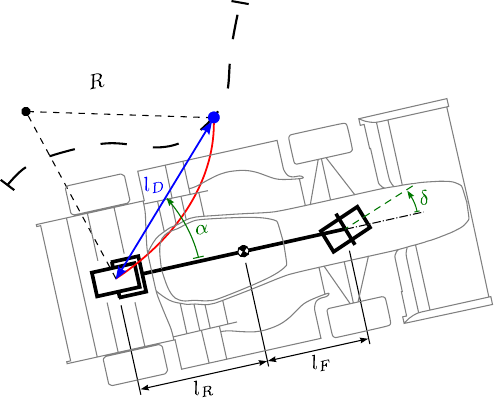}
                \caption{Shows the geometric relationship between the vehicle and the desired trajectory.}
                \label{fig:pupu}
            \end{minipage}
            \hfill
            \begin{minipage}{4cm}
                \centering
                \includegraphics[width=4cm]{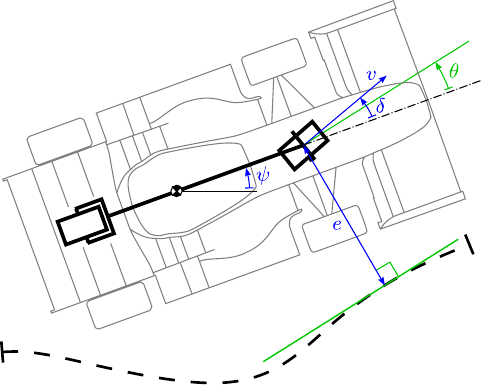}
                \caption{The kinematic model for a front-steering vehicle represented in the trajectory frame.}
                \label{fig:stanley}
            \end{minipage}
        \end{figure}

    Initially, we relied solely on a lateral control algorithm, the Stanley controller \cite{stanley_controller}, first used on off-road terrain proposed in \cite{stanley_controller}. The said control strategy is based on a kinematic vehicle model shown in \hyperref[fig:stanley]{Figure \ref{fig:stanley}}, and the steering law is
        \begin{equation}
            \delta_\mathrm{st} = \arctan\left(\frac{k e}{k_\mathrm{soft} + v}\right) + \left(\theta - \theta_\mathrm{ss}\right),
        \end{equation}
        where $e$ is the cross-track error, $v$ is the velocity, $\theta$ is the relative yaw angle, $k_\mathrm{soft}$ is tuned to avoid oversensitivity at low speeds and kept $\theta_\mathrm{ss}$, the steady state yaw introduced in \cite{stanley_controller} which is given by 
        \begin{equation}
		\theta_\mathrm{ss} = \frac{m r_\mathrm{traj} v}{C_y \left(1+\frac{l_\mathrm{F}}{l_\mathrm{R}}\right)}
        \end{equation}
        where $m$ is the car's mass, $C_y$ is the tire stiffness on the front axle, and $r_\mathrm{traj}$ is the yaw rate of the trajectory. $l_\mathrm{F}$ and $l_\mathrm{R}$ are the distances between the vehicle's centre of gravity and the front or rear axis. Retaining $\theta_\mathrm{ss}$ significantly improved the overall performance of the controller on the complex track layouts that we have to tackle by making the car point more inward when cornering. However, due to the inherent characteristics of the control policy and its attempt to follow the centerline -- often resulting in a trajectory that may not be feasible to track -- a substantial amount of overshoot was still observed in tight and sharp corners.

        To address these issues, we implemented a method to combine the output of the Stanley controller with those of the Pure Pursuit controller, similar to the approach in \cite{combined_controller}. This combination does not compromise all of the global asymptotic stability of the Stanley controller, as proven in \cite{stanley_controller}, because the Pure Pursuit controller is also stable, with its stability conditions analysed in \cite{Murphy1994} and \cite{PuPuStability}. We adopted the following formulation
        \begin{equation}
		  \delta_\mathrm{pp} = \arctan\left(\frac{2\left(l_\mathrm{F}+l_\mathrm{R}\right)\sin\alpha}{L_\mathrm{d}}\right)
        \end{equation}
        where $l_\mathrm{F}+l_\mathrm{R}$ represents the wheelbase and $\alpha$ denotes the angular offset. The look-ahead distance, $L_\mathrm{d}$, is updated at each iteration based on the current velocity, as shown in
        \begin{equation}
            L_\mathrm{d} = L_\mathrm{d,min} + k_v v.
        \end{equation}
        with $L_\mathrm{d,min}$ being the minimum look-ahead distance and $k_v$ being the velocity gain constant. Both $L_\mathrm{d,min}$ and $k_v$ are tuning parameters. This modification is practical at higher velocities, as the controller becomes less aggressive and can initiate cornering earlier. Furthermore, an advantage of this approach over the Stanley controller is that it does not strictly attempt to track the trajectory; instead, it aims to steer toward the control point in a geometrically feasible manner, as shown in \hyperref[fig:pupu]{Figure \ref{fig:pupu}}.
        
        We then compute the linear combination of both controllers
        \begin{align}
            \delta_\mathrm{c} &= k_\mathrm{st} \delta_\mathrm{st} + k_\mathrm{pp} \delta_\mathrm{pp} 
            \\
            1 &= k_\mathrm{st} + k_\mathrm{pp} & 0 \le k_\mathrm{st} \text{, } k_\mathrm{pp}
        \end{align}
        where $k_\mathrm{pp}$ is calculated as a function of the curvature, $\kappa$, at the look-ahead distance
        \begin{equation}
            k_\mathrm{pp} = \min \left( k_\mathrm{min} + \frac{|\kappa|}{\kappa_\mathrm{ref}} k_\mathrm{curve} , k_\mathrm{max} \right)
        \end{equation}
        $k_\mathrm{min}$, $k_\mathrm{max}$, $\kappa_\mathrm{ref}$, and $k_\mathrm{curve}$ can be selected after assessing the scenarios each control scheme shows superior results. This approach allows us to rely more on the Pure Pursuit controller when approaching corners and shift towards the Stanley controller when exiting, enabling faster recovery.

        The final modification we introduce, \hyperref[eq:yaw]{Equation \ref{eq:yaw}}, compensates for the tyre forces by modelling them as active damping, following the approach used in \cite{stanley_controller}. This adjustment was essential, allowing us to utilise the torque vectoring system that was already available in manual mode.
        \begin{equation}
            k_\mathrm{d,yaw} \left( r_\mathrm{meas} - r_\mathrm{traj} \right)
            \label{eq:yaw}
        \end{equation}

    The lateral control schemes were evaluated in simulation on multiple tracks, with performance metrics summarised in \hyperref[tab:controller_comparison]{Table \ref{tab:controller_comparison}}. The results demonstrate that the Stanley controller achieves accurate path tracking under ideal conditions, but its strict adherence to the centerline often results in significant overshoot, which is even more pronounced during on-track runs. In contrast, the Pure Pursuit controller produces increased tracking errors, but achieves better lap times due to the behaviour already discussed. The combined controller leverages the strengths of both approaches, allowing it to adaptively select the most suitable behaviour in each scenario, resulting in improved recovery characteristics, faster lap times, and enhanced overall robustness.

    \begin{table}[ht]
        \centering
        \setlength{\tabcolsep}{4pt}

        \caption{Performance comparison of controllers on the track used in the 2023 Formula Student Germany competition}
        
        \begin{tabular}{lccccc}
            \hline
            & \textbf{Combined} & \textbf{Stanley} & \textbf{Pure Pursuit} \\
            \hline
            ITAE & 143.0 & 50.0 & 273.0 \\
            RMS [m] & 0.201 & 0.075 & 0.374 \\
            Max lateral error [m] & 0.645 & 0.244 & 1.047 \\
            Lap time [s] & 40.772 & 41.730 & 41.248 \\
            \hline
        \end{tabular}
        \label{tab:controller_comparison}
    \end{table}

    \subsubsection{Longitudinal Control}

        We use a simple P controller to control the throttle and brake actions. The difference between estimated and planned velocities drives the P controller, which converts this error signal into an appropriate torque demand. The resulting torque demand is then forwarded to the low-level controllers, which manage the vehicle's electric motors for throttle and braking actions. Although the P controller offers simplicity and effective performance in most cases, steady-state errors arise in certain scenarios. To address this, incorporating an integral term could enhance the accuracy and robustness of the system without significantly increasing complexity. 
        
\section{Results}
\label{sec:results}

This development marks the end of an unsuccessful period of driverless results for the team, which is becoming increasingly important as competition rules place more and more emphasis on autonomous technologies. For the first

    \begin{figure}[!ht]
     \centering
     \includegraphics[width=8.0cm]{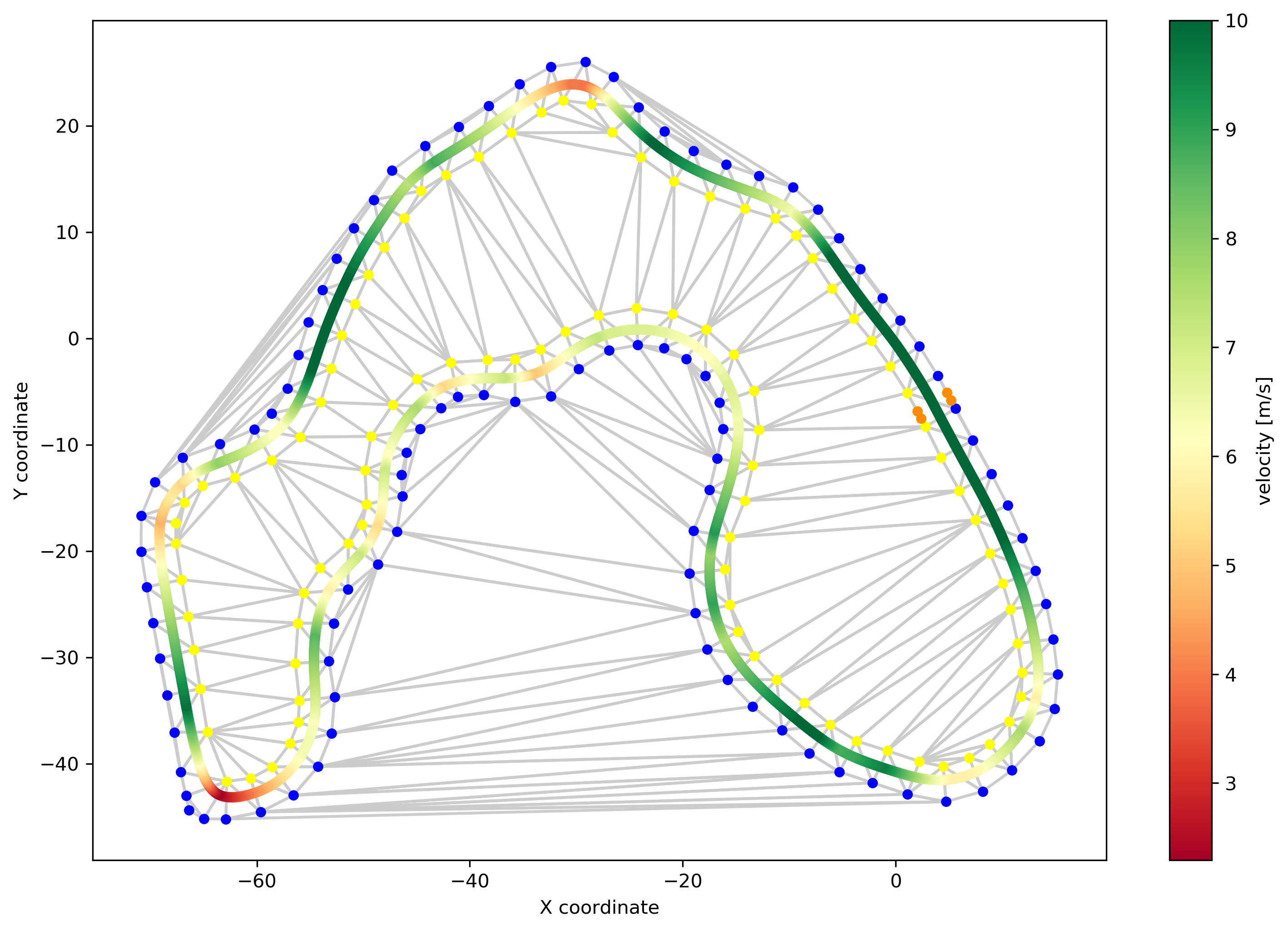}
     \caption{The planned path, where circles represent cones bounding the track, the triangulation is marked by gray lines, and the final trajectory is shown by the spline colored by target velocity.}
      \label{fig:combined_traj}
    \end{figure}
\begin{figure}[!ht]
     \centering
     \includegraphics[width=8.0cm]{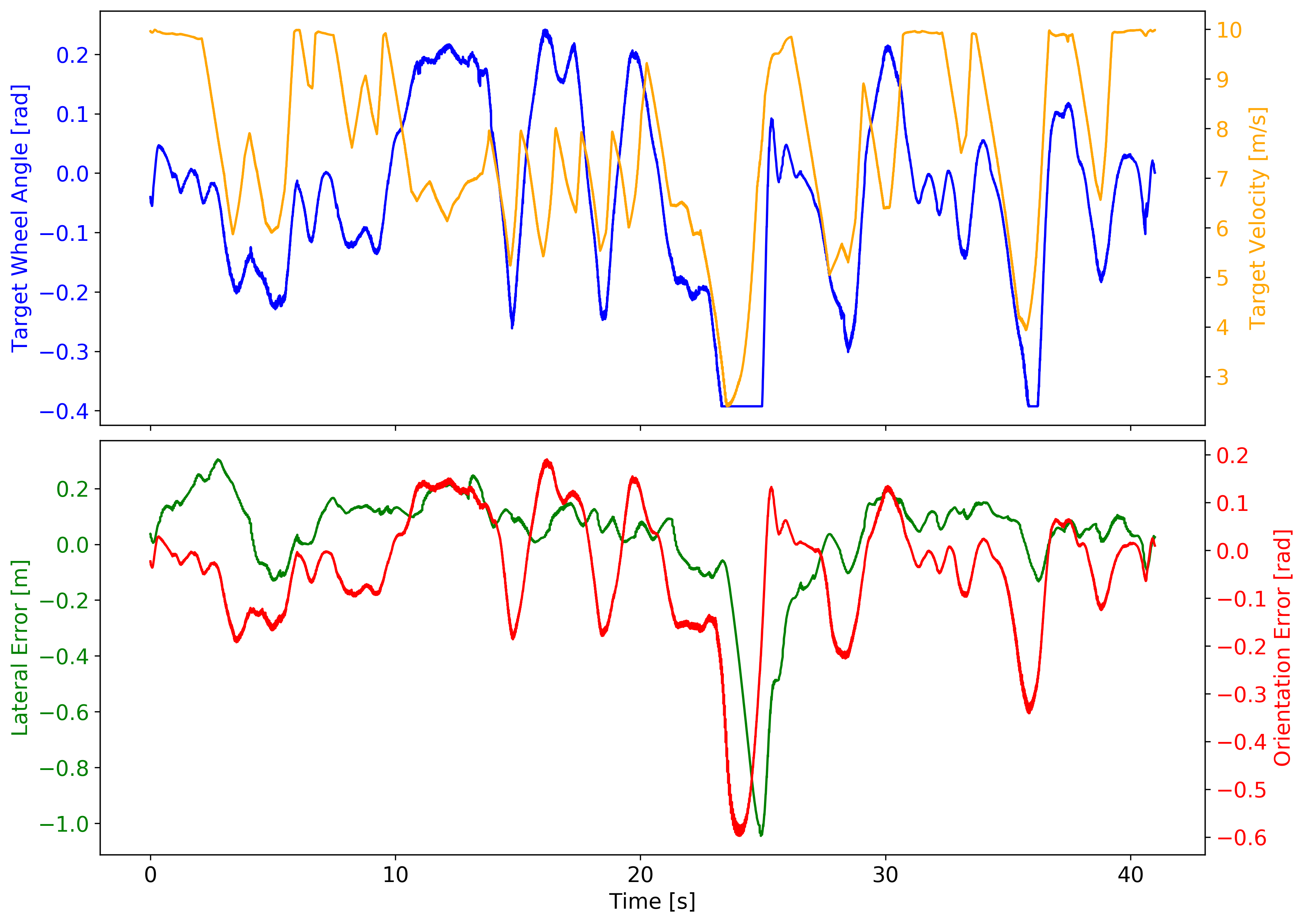}
     \caption{Controller data, showing the target wheel angle and velocity, and the deviation from the reference trajectory by lateral and orientation error.}
      \label{fig:control_data}
    \end{figure}


 \noindent time in the team's history, Formula Student East and the overall third place in the electric-driverless category at Formula Student Germany (FSG). In 2023 at FSG, the team also got Formula Student East and the overall third place in the electric-driverless category at Formula Student Germany (FSG). In 2023 at FSG, the team also got its first podium finish in the trackdrive event with a third place and achieved several honorable mention top five finishes. At the trackdrive event, the average lap time completed by our autonomous system was $41.30s$, with its fastest lap being $41.02s$. The path planning result for the fastest lap is shown in \hyperref[fig:combined_traj]{Figure \ref{fig:combined_traj}}, and the corresponding path tracking data and error terms are shown in \hyperref[fig:control_data]{Figure \ref{fig:control_data}}.

\section{Conclusions}
\label{sec:conclusions}

In this article, we discuss the autonomous software stack and architecture developed by BME Formula Racing Team within a 5-year time frame for the FSD series. We detail our approaches with respect to state estimation, perception, planning, and control in hope of providing a comprehensive framework on which other projects can build on. The practical application of the presented methods is supported by the many kilometres of testing and the results achieved in prestigious races. Future work will focus on improving the robustness of perception, implementing a more extensive state estimation, and exploring non-linear control methods.

The expertise gained from this project plays a key role in participating in high-profile initiatives, such as A2RL, which demonstrates that STEM education and university competitions help bridge the gap between academic research and real-world applications. By sharing our approach to driverless racing, we want to contribute to the autonomous racing community and inspire other participating teams to share their valuable insight on the topic.


\addtolength{\textheight}{-12cm}

\section*{Acknowledgement}
\label{sec:acknowledgement}
It takes the hard work and dedication of many people to make a formula student project a reality, and for that we are grateful. We would like to thank all current and former team members of BME Formula Racing Team who have contributed to the achievements of the driverless group. In addition, we express our sincere appreciation to all sponsors of the team for their unwavering support. Their continued backing allows us to pursue our passion for racing.


\bibliographystyle{IEEEtran}
\bibliography{references}

\end{document}